\documentclass{WileyMSP-template}

\usepackage{textcomp}
\usepackage{graphicx}
\usepackage{float}
\usepackage[english]{babel}
\usepackage{amssymb}
\usepackage[normalem]{ulem}
\usepackage{amsmath}
\usepackage{amsthm}
\usepackage{amsfonts}
\usepackage{enumerate}
\usepackage{mathtools}
\usepackage{hyperref} 
\usepackage[dvipsnames]{xcolor}

\usepackage{lineno}
\usepackage{bm}

\usepackage{algorithmic}
\usepackage{algorithm}
\usepackage{ragged2e}
\usepackage{cite}

\begin{document}

\pagestyle{fancy}
\rhead{\includegraphics[width=2.5cm]{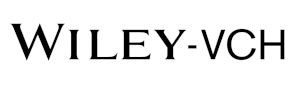}}

\title{Harnessing Discrete Differential Geometry: A Virtual Playground for the Bilayer Soft Robotics}

\maketitle


\author{Jiahao Li$^{1}$}
\author{Dezhong Tong$^{1}$}
\author{Zhuonan Hao$^{1}$}
\author{Yinbo Zhu}
\author{Hengan Wu}
\author{Mingchao Liu$^*$}
\author{Weicheng Huang$^*$}

\begin{affiliations}

$^{1}$ Equal contribution.\\
$*$ Corresponding authors. \\

\hfill \break

Mr. Jiahao Li, Prof. Yinbo Zhu, Prof. Hengan Wu \\

CAS Key Laboratory of Mechanical Behavior and Design of Materials, Department of Modern Mechanics, University of Science and Technology of China, Hefei 230027, People’s Republic of China

Dr. Dezhong Tong \\
Department of Material Science and Engineering, University of Michigan, Ann Arbor, Michigan, 48105, USA

Mr. Zhuonan Hao \\
Department of Mechanical and Aerospace Engineering, University of California, Los Angeles,  California, 90095, USA

Prof. Mingchao Liu \\
Department of Mechanical Engineering, University of Birmingham, Birmingham, B15 2TT, UK\\
m.liu.2@bham.ac.uk (Corresponding author)

Prof. Weicheng Huang\\
School of Engineering, Newcastle University, Newcastle upon Tyne, NE1 7RU, UK\\
weicheng.huang@newcastle.ac.uk (Corresponding author)

\end{affiliations}


\keywords{Soft robotics, Strain mismatch, Bilayer system, Discrete elastic rod}

\begin{abstract}

\begin{justify}

Soft robots have garnered significant attention due to their promising applications across various domains.
A hallmark of these systems is their bilayer structure, where strain mismatch caused by differential expansion between layers induces complex deformations.
Despite progress in theoretical modeling and numerical simulation, accurately capturing their dynamic behavior, especially during environmental interactions, remains challenging.
This study presents a novel simulation environment based on the Discrete Elastic Rod (DER) model to address the challenge.
By leveraging discrete differential geometry (DDG), the DER approach offers superior convergence compared to conventional methods like Finite Element Method (FEM), particularly in handling contact interactions -- an essential aspect of soft robot dynamics in real-world scenarios.
Our simulation framework incorporates key features of bilayer structures, including stretching, bending, twisting, and inter-layer coupling.
This enables the exploration of a wide range of dynamic behaviors for bilayer soft robots, such as gripping, crawling, jumping, and swimming.
The insights gained from this work provide a robust foundation for the design and control of advanced bilayer soft robotic systems.

\end{justify}

\end{abstract}


\begin{justify}

\section{Introduction}
Robotics is the science of designing and constructing machines capable of movement, perception, and cognition to assist humans in performing various tasks. Inspired by living organisms, using soft matter in robot design has gained significant attention in recent decades. The inherent compliance of soft bodies allows them to adapt to complex environments, enabling innovative applications in fields such as healthcare, agriculture, and the food industry~\cite{wang2019soft,shin2018hygrobot,li2022dual,xu2022insect,hu2017electrically,yin2021visible,huang2022design,tao2021morphing,kim2022magnetic,yang2023morphing}. Given the potential of soft robots, various functional materials, such as liquid crystal elastomers, pneumatic actuators, and light-driven systems, have been explored as actuators due to their ability to deform in response to diverse external stimuli. However, the intrinsic compliance and nonlinearity of soft materials pose significant challenges in achieving precise and effective deformation control, which limits their practical effectiveness in real-world applications.

A widely adopted approach to addressing this challenge is using bilayer structures in soft robot design. Inspired by natural phenomena such as the opening of pea pods, a bilayer structure consists of two layers—an top and a bottom layer—adhered at their interface~\cite{armon2011geometry}, as illustrated in Fig.~\ref{fig:overview}A.
When one layer undergoes expansion, a mismatch strain arises at the interface. This strain enables programmable deformations, including bending, twisting, or curling, in response to external stimuli such as temperature, moisture, and light. Such design versatility makes bilayers particularly appealing for tasks requiring complex motions, such as crawling (Fig.~\ref{fig:overview}B), jumping (Fig.~\ref{fig:overview}C), and swimming (Fig.~\ref{fig:overview}D). The widespread use of bilayers in soft robotics highlights their practicality and potential for creating efficient, multifunctional devices. However, a comprehensive understanding of the underlying mechanisms that govern bilayer behavior remains incomplete.

\begin{figure}
  \centering  \includegraphics[width=0.9\linewidth]{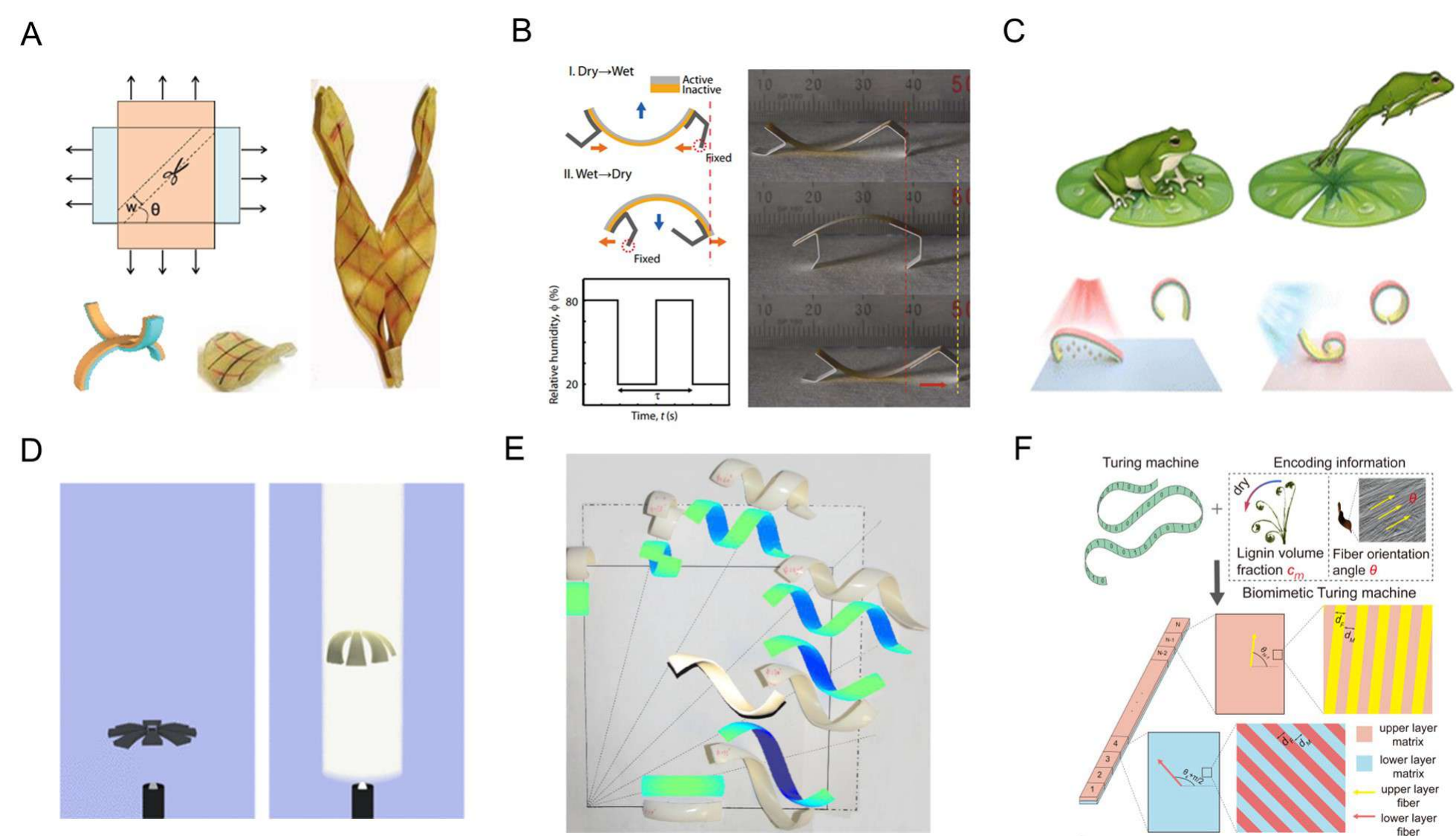}
  \caption{Examples of bilayer soft robots and structures. (A) The helix bilayer hydrogel inspired from the chiral pea pod~\cite{armon2011geometry}. (B) Moisture-driven crawling robot consisting of a hygroscopically active layer and a hygroscopically inactive layer~\cite{shin2018hygrobot}. (C) Dual-responsive jumping actuator driven by light and humidity, composed of a photothermal expansion layer, an interfacial adhesion layer, and a moisture-responsive layer~\cite{li2022dual}. (D) Light-driven jellyﬁsh-like swimming robot fabricated from the striped composite hydrogel~\cite{yin2021visible}. (E) Tunable helical ribbon fabricated by bonding an unstrained elastic adhesive sheet to a pre-strained latex sheet~\cite{ChenZi2010}. (F) Inverse design of the target space curve by encoding the microstructure of the morphing bilayer ribbons~\cite{lijh2025}. }
  \label{fig:overview}
\end{figure}

The theoretical foundation of bilayer structures dates back to 1925, when Timoshenko~\cite{Timoshenko1925} first explored the bi-metal thermostats. Building on this foundational work, Chen et al.~\cite{ChenZi2010} extended the model to include three-dimensional (3D) deformations, capturing morphologies such as helix and helicoids, as illustrated in Fig.~\ref{fig:overview}E. However, their model was limited to illustrate the morphing bilayer ribbon with constant curvature and torsion, neglecting the fiber-matrix microstructures, which leads to the anisotropic mismatch strain. Recently, Li et al.~\cite{lijh2025} developed a general theory for the inverse design of target space curves with varying curvature and torsion along the arc length parameter, as shown in Fig.~\ref{fig:overview}F. This multiscale theoretical framework bridges the gap between microstructural properties and macroscopic morphologies in morphing bilayer ribbons.

Although significant progress has been made in the theoretical development of bilayer structures, these theories are often based on simplifying assumptions, limiting their applicability for studying soft robots constructed using bilayer structures. This limitation underscores the urgent need for developing robust numerical simulations tailored to bilayer systems. Existing numerical methods for simulating bilayer structures can be broadly categorized into two main approaches. The first involves reformulating the bilayer system into an equivalent monolayer representation, allowing the elastic energies of the system to be expressed as the $L_2$ norm of the first and second fundamental forms between the initial configuration and the intermediate configuration~\cite{Efrati2009}. 
While this approach provides a simplified solution, it introduces additional energy formulations directly tied to specific actuation mechanisms. This limitation hinders the development of a general numerical framework applicable to diverse soft robot designs and lacks a clear physical interpretation, reducing its generality for broader applications~\cite{VanRees2017}. 
Another approach employs the finite element method (FEM) to simulate both layers simultaneously by incorporating constraints at their interface~\cite{VanRees2017, lijh2025}. However, the design of soft robots is often highly intricate, and directly using FEM software can result in low computational efficiency. This inefficiency poses challenges for optimizing designs and exploring sim-to-real methods for soft robots.

In recent years, discrete differential geometry (DDG)-based simulations have attracted significant interest within the soft robotics community due to their computational efficiency, achieved through reduced-order expressions, and their numerical robustness in handling nonlinear deformations, attributed to precise geometric representations~\cite{grinspun2006discrete}. The physical accuracy of DDG-based simulations has been validated in various applications, including buckling~\cite{tong2023snap, tong2021automated}, frictional contact~\cite{choi2021implicit, tong2023fully}, and robotic manipulation~\cite{choi2023deep, tong2024sim2real}. 
These successes have led to the adoption of such simulations in modeling a variety of soft robots, including shape-memory alloy (SMA)-actuated robots~\cite{huang2020dynamic}, magnetic soft robots~\cite{huang2023modeling}, flagella-inspired robots~\cite{huang2020numerical}, jumping robots~\cite{tong2024inverse}, and space net robots~\cite{huang2024dynamic}.
More recently, a general soft robot simulator has been developed using DDG-based simulations~\cite{choi2024dismech}. However, these simulations often overlook the actuation mechanisms of soft robots, which are frequently based on bilayer structures. In many cases, the mentioned prior works measured the deformations of the robots experimentally and then fed them to simulations, bypassing the complex interactions within multilayer structures. Moreover, the absence of a clear physical interpretation of bilayer structures makes it difficult to connect different actuation mechanisms to the resulting deformations. This gap prevents the development of unified, predictive models, limiting the potential for optimizing and generalizing soft robot designs.

In this paper, we introduce a general numerical framework for simulating soft robots actuated by bilayer structures to uncover the intrinsic interactions between intelligent soft materials and robotic systems. Building on the discrete differential geometry framework, we propose a novel internal energy formulation based on the geometry of the soft robot, ensuring non-translation and non-rotation constraints at the bilayer interface. Additionally, we have developed and released a general soft robot simulator specifically tailored for modeling soft robots actuated by bilayer structures. \footnote{See \href{https://github.com/DezhongT/Bilayer_Soft_Robots_Sim}{https://github.com/DezhongT/Bilayer\_Soft\_Robots\_Sim}}
To demonstrate the capabilities of this simulation environment, a series of examples, including gripping, crawling, jumping, and swimming are presented to showcase its potential for diverse soft robotic applications.

This paper is organized as follows. 
In Sec.~\ref{sec:simulation}, we detail the modeling procedure of the bilayer slender system.
Next, in Sec.~\ref{sec:validation}, we validate our numerical framework by using classical examples.
Moving forward, in Sec.~\ref{sec:demo}, several representative scenarios from previous studies are simulated.
Finally, in Sec.~\ref{sec:conclusion}, we summarize our primary contributions and identify promising avenues for future research.
A simplified 2D model for a planar bilayer beam system is discussed in Appendix A.

\section{Simulation Environment}
\label{sec:simulation}

In this section, the numerical model for the nonlinear dynamics of a flexible bilayer rod is formulated.
The DDG-based framework for a single elastic rod is first introduced~\cite{Khalid2008, BergouDVT, bergou2008discrete}, followed by the stick coupling between two rods at the interface.
Finally, the time-marching method for the solving of nonlinear equations of motion is discussed~\cite{Huang2019}.
Here, we consider a slender body with length $L$, width $w$, thickness $h$, and is assumed to be manufactured by isotropic and linearly elastic material.
The top layer is of Young's modulus $E_{1}$, shear modulus $G_{1}$, and thickness $h_{1}$, and the bottom layer is of Young's modulus $E_{2}$, shear modulus $G_{2}$, and thickness $h_{2}$.
Note that $h$ is the thickness of the bilayers and $h= h_{1} + h_{2}$.
We use the centerline of each layer to describe the configuration of the bilayer system, i.e., the centerline of the top layer is denoted as $s_{1}$, and the centerline of the bottom layer is denoted as $s_{2}$ as shown in Fig.~\ref{fig:model}A. 

\begin{figure}
  \centering  \includegraphics[width=0.8\linewidth]{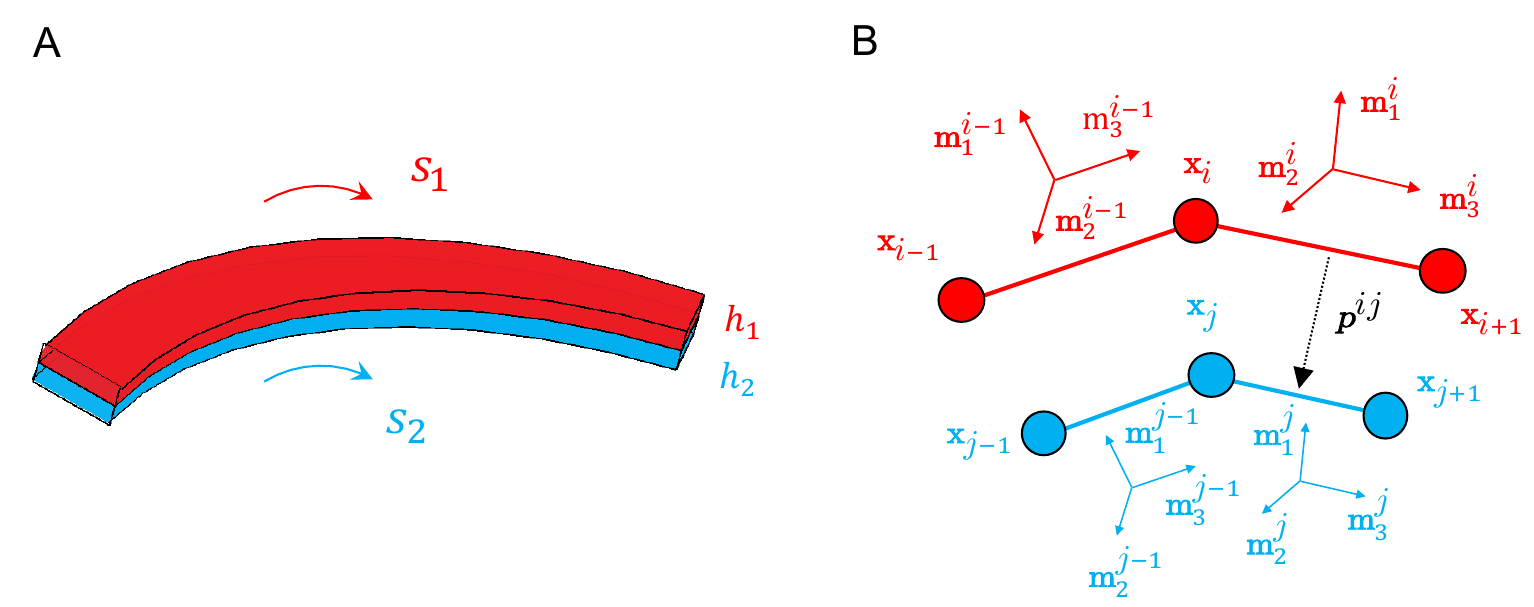}
  \caption{Schematic representation of a bilayer structure. (A) Diagram of a bilayer system that composed of an top layer (red) and bottom layer (blue), with arc length ${s}_1$ and ${s}_2$, and thickness ${h}_1$ and ${h}_2$, respectively. (B) Discretization diagram of the bilayer structure. The subscripts and superscripts $i$ and $j$ denote the discrete element indices of top layer and bottom layer, respectively, e.g., node position $\mathbf{x}$, edge material frame $\{\mathbf{m}^{i}_{1}, \mathbf{m}^{i}_{2}, \mathbf{m}^{i}_{3}\}$ and the relative position $\mathbf{p}^{ij}$ between top layer and bottom layer. }
  \label{fig:model}
\end{figure}

\paragraph{Degrees of freedom} We use Kirchhoff model to capture the geometrically nonlinear deformation of each layer.
To capture its geometrically nonlinear configuration, the centerline of two layers is discretized into $N$ and $M$ nodes, respectively.
We use $i \in [0, N-1]$ index to represent the quantity associated with the top layer and use $j \in [0, M-1]$ index to represent the quantity associated with the bottom layer, and the total degrees of freedom (DOF) vector is denoted as $\mathbf{q} \in \mathbb{R}^{\mathcal{N} \times 1}$, where $\mathcal{N}$ is the total DOF number. 
Here, we formulate the elastic energy of the top layer, and the same formulation can be used for the bottom layer.
The total elastic energies of a deformed top layer are comprised of three parts: stretching, bending, and twisting~\cite{audoly2000elasticity}.

\paragraph{Stretching element} The stretching element is associated with two consecutive nodes,
\begin{equation}
\left\{ \mathbf{x}_{i}, \mathbf{x}_{i+1} \right\} \in \mathbb{R}^{6 \times 1},
\end{equation}
and the edge vector is given by
\begin{equation}
\mathbf{e}^{i} = \mathbf{x}_{i+1} - \mathbf{x}_{i}.
\end{equation}
Its uniaxial strain is 
\begin{equation}
\epsilon^{i} = \frac{|| \mathbf e^{i} ||}{|| \bar{\mathbf{e}}^{i} ||} - 1.
\label{eq:StrechingStain}
\end{equation}
To follow the convention, a bar on top indicates the quantities evaluated in the undeformed configuration.
The linear elastic stretching energy is a quadratic function of uniaxial strain,
\begin{equation}
\mathcal{S}^{i} = \frac {1} {2} EA (\epsilon^{i})^2 || \bar{\mathbf{e}}^{i} || 
\end{equation}
where $EA$ is the local stretching stiffness.

\paragraph{Bending and twisting element} Moving forward, we discuss the bending and twisting element of a rod.
As shown in Fig.~\ref{fig:model}B, to formulate the elastic bending and twisting energies, two special frames are introduced.
Each edge has an orthonormal reference frame,
\begin{equation}
\left\{\mathbf{d}^{i}_{1}, \mathbf{d}^{i}_{2}, \mathbf{d}^{i}_{3} \right\},
\end{equation}
and a material frame,
\begin{equation}
\left\{\mathbf{m}^{i}_{1}, \mathbf{m}^{i}_{2}, \mathbf{m}^{i}_{3}\right\}.
\end{equation}
Both the frames share the tangent as one of the directors,
\begin{equation}
\mathbf{d}^{i}_{3} \equiv \mathbf{m}^{i}_{3} = \frac {\mathbf e^{i}} {|| \mathbf e^{i} ||},
\end{equation}
and the difference between two frames is captured by a scalar called rotational angle, $\theta^{i}$, 
\begin{equation}
\begin{aligned}
\mathbf{m}^{i}_{1} &= \mathbf{d}^{i}_{1} \cos \theta^{i} + \mathbf{d}^{i}_{2} \sin \theta^{i}, \\
\mathbf{m}^{i}_{2} &= \mathbf{d}^{i}_{2}  \cos \theta^{i} - \mathbf{d}^{i}_{1}\sin \theta^{i}.
\end{aligned}
\end{equation}
The bending and twisting elements are associated with three nodes and two rotational angles,
\begin{equation}
\left\{ \mathbf{x}_{i-1}, \theta^{i-1}, \mathbf{x}_{i}, \theta^{i}, \mathbf{x}_{i+1} \right\} \in \mathbb{R}^{11 \times 1 }.
\end{equation}
The normalized bending strain is based on the curvature binormal that measures the misalignment between two consecutive edges,
\begin{equation}
(\mathbf{\kappa b})_{i} = \frac {2 \left (\mathbf{e}^{i-1} \times \mathbf{e}^{i} \right) } { || \mathbf{e}^{i-1} || \; || \mathbf{e}^{i} || + \mathbf{e}^{i-1} \cdot \mathbf{e}^{i} }.
\end{equation}
Next, the two material curvatures $\{\eta_{i},  \xi_{i} \}$, are given by the inner products between the curvature binormal and material frame vectors,
\begin{equation}
\begin{aligned}
\chi_{i} & = \frac{1}{2} \frac {\left( \mathbf m_2^{i-1} + \mathbf m_2^{i} \right)} {\Delta \bar {l}_{i}} \cdot (\kappa \mathbf b)_{i}, \\
\xi_{i} & = - \frac{1}{2} \frac {\left( \mathbf m_1^{i-1} + \mathbf m_1^{i} \right)} {\Delta \bar {l}_{i}} \cdot (\kappa \mathbf b)_{i}.
\end{aligned}
\label{eq:bendC} 
\end{equation}
where ${\Delta \bar {l}}_{i} = (|| \bar{\mathbf{e}}^{i-1} || + || \bar{\mathbf{e}}^{i} || ) /2$ is its Voronoi length.
The twisting curvature at the $i$-th node, in the discrete setting of the discrete rod model, is measured by using the discrete twist
\begin{equation}
\tau_{i} = \frac {(\theta_{i} - \theta_{i-1} + {\theta}^{\mathrm{ref}}_{i} )} { \Delta \bar {l}_{i} },
\label{eq:twistC} 
\end{equation}
where $ {\theta}^{\mathrm{ref}}_{i} $ is the reference twist associated with the reference frame~\cite{jawed2018primer}.
Similarly, the bending energy as well as the twisting energy associated with $(i)$-th element is
\begin{equation}
\mathcal{B}_{i} = \frac {1} {2}  {EI_{1}} (\chi_{i} - \bar{\chi}_{i})^2  { {\Delta \bar{l}}_{i} }  + \frac {1} {2}  {EI_{2}}  (\xi_{i} - \bar{\xi}_{i})^2 { {\Delta \bar{l}}_{i} } + \frac {1} {2}  {GJ} ( \tau_{i} - \bar{\tau}_{i})^2 { {\Delta \bar {l}}_{i} } ,
\end{equation}
where $EI_{1}$, $EI_{2}$, and $GJ$ are the local stiffness parameters.
The total elastic energy for the top layer is the sum of the stretching, bending, and twisting,
\begin{equation}
\mathcal{E}_{1} = \sum_{i=0}^{N-1} \mathcal{S}^{i} + \sum_{i=1}^{N-2} \mathcal{B}_{i}.
\end{equation}
The formulation for the bottom layer is similar to the top layer,
\begin{equation}
\mathcal{E}_{2} = \sum_{j=0}^{M-1} \mathcal{S}^{j} + \sum_{j=1}^{M-2} \mathcal{B}_{j}.
\end{equation}

\paragraph{Coupling element} Next, we discuss the coupling between the two layers at the interface.
The discrete model considers the space curve as the points in 3D space and the local directors (the material frame) at each point. 
Therefore, we need to couple both the node position and the material frame of the two rods.
The coupling element is associated with $i$-th edge and $j$-th edge (four nodes and two angles),
\begin{equation}
\left\{ \mathbf{x}_{i}, \theta^{i}, \mathbf{x}_{i+1}, \mathbf{x}_{j}, \theta^{j}, \mathbf{x}_{j+1} \right\} \in \mathbb{R}^{14 \times 1}.
\end{equation}
The Voronoi point of each edge for the bilayers can be defined as:
\begin{equation}
\begin{aligned}
\mathbf{p}^{i} &= \frac {1} {2} (\mathbf{x}_{i} + \mathbf{x}_{i+1}), \\
\mathbf{p}^{j} &= \frac {1} {2} (\mathbf{x}_{j} + \mathbf{x}_{j+1}),
\end{aligned}
\end{equation}
thus the relative displacement as shown in Fig.~\ref{fig:model} can be defined as:
\begin{equation}
\mathbf{p}^{ij} = \mathbf{p}^{i} - \mathbf{p}^{j}.
\end{equation}
To achieve the non-translation condition between the two edges at the interface, we project the relative displacement between the two edges, $\mathbf{p}^{ij}$, onto the material frame in three directions, and use a penalty energy method to constrain it,
\begin{equation}
\begin{aligned}
&\mathcal{C}^{ij}_{T} = \frac{1}{2} K_{T} \left[  \left(  \mathbf{p}^{ij}  \cdot {\mathbf{m}_1^{ij}} - {h} \right) ^2 + \left(  \mathbf{p}^{ij}  \cdot {\mathbf{m}_2^{ij}} \right) ^2 +  \left(  \mathbf{p}^{ij}  \cdot {\mathbf{m}_3^{ij}} \right) ^2 \right], \\
& \mathrm{with} \; \mathbf{m}_{\alpha}^{ij} = \frac {1} {2}(\mathbf{m}_{\alpha}^{i}  + \mathbf{m}_{\alpha}^{j} ), \; \alpha \in \{1,2,3 \},
\end{aligned}
\end{equation}
where  $K_{T}$ is a penalty stiffness.
Then, the total coupling energy for translation can be written as the sum of all interface edges,
\begin{equation}
\mathcal{C}_{T} = \sum_{\substack{ij}}^{K} \mathcal{C}^{ij}_{T},
\end{equation}
where $K$ is the total number of coupling interfaces.
Next, the coupling between the two material frames is considered to achieve the non-rotation condition between the two rod segments.
This can be achieved by using the classical bending and twisting elements in the  DER method, i.e., the relative bending and twisting between $i$-th edge and $j$-th edge is relatively large and fixed as zeros. 
The bending and twisting coupling between $i$-th edge and $j$-th edge is given by
\begin{equation}
\mathcal{C}^{ij}_{R} = \frac {1} {2}  {K}_{R} \left[ (\chi^{ij} - \bar{\chi}^{ij} )^2 + (\xi^{ij} - \bar{\xi}^{ij})^2  + (\tau^{ij} - \bar{\tau}^{ij} )^2\right] {\Delta \bar{l}}^{ij} ,
\end{equation}
where $\chi$, $\xi$, $\tau$, are the relative bending and twisting curvatures between the $i$-th edge and the $j$-th edge and can be computed in a way similar to Eq.~(\ref{eq:bendC}) and Eq.~(\ref{eq:twistC}), $K_{R}$ is the penalty stiffness, and ${\Delta \bar {l}}^{ij} = (|| \bar{\mathbf{e}}^{i} || + || \bar{\mathbf{e}}^{j} || ) /2$ is the average length of the two edges. Finally,  the total coupling energy for rotation is the sum of all interface edges,
\begin{equation}
\mathcal{C}_{R} = \sum_{\substack{ij}}^{K} \mathcal{C}^{ij}_{R}.
\end{equation}
The total potential energy of a bilayer system is defined as
\begin{equation}
\mathcal{E} = \mathcal{E}_{1} + \mathcal{E}_{2} + \mathcal{C}_{T} + \mathcal{C}_{R}. 
\end{equation}

\paragraph{Friotnal contact force} For the simulation of real robotic applications, i.e., gripping and crawling, the numerical model needs to take the frictional contact force into account.
We here use the incremental potential method for the contact simulation.
When a node $\mathbf{x}_{i} \in \mathbb{R}^{3 \times 1}$ gets into contact with the object, the bilayer system is subjected to a contact force.
A smooth log-barrier reaction force is considered when a node is close to the object ~\cite{li2020incremental,li2020codimensional},
\begin{equation}
(\mathbf{F}^{\mathrm{con}}_{i})^{n} =
\begin{cases}
K_{c} \left[ - 2 (d_{i} - \hat{d}) \log( \frac {d_{i}} {\hat{d}}) - \frac {(d_{i} - \hat{d})^2} {d_{i}} \right] \mathbf{n}_{i} \; &\mathrm{when} \; 0 < d_{i} < \hat{d}, \\
\mathbf{0} \; & \mathrm{when} \; d_{i} \geq \hat{d},
\end{cases}
\label{eq:barrierContact}
\end{equation}
where $\mathbf{n}_{i}$ is the normal direction of the contact surface, $d_{i}$ is the contact distance between the node and the target, $\hat{d}$ is the barrier parameter, and $K_{c}$ is the contact stiffness.
Moreover, when the velocity of a contact node along the surface tangential direction is non-zero, i.e., 
\begin{equation}
\mathbf{v}_{i}^{t} =\mathbf{v}_{i} - \mathbf{v}_{i} \cdot \mathbf{n}_{i} \neq \mathbf{0},
\end{equation}
the frictional force contributes to the motion of the dynamical system.
Here, a smooth frictional force is employed based on the maximum dissipative
principle ~\cite{li2020incremental,li2020codimensional},
\begin{equation}
(\mathbf{F}^{\mathrm{con}}_{i})^{t} =
\begin{cases}
- \mu \; || (\mathbf{F}^{\mathrm{con}}_{i})^{n} || \; \frac {\mathbf{v}_{i}^{t}} {||\mathbf{v}_{i}^{t}||} \left(- \frac {||\mathbf{v}_{i}^{t}||^2} {\epsilon_{v}^2} +  \frac {||\mathbf{v}_{i}^{t}||} {\epsilon_{v}} \right) \; &\mathrm{when} \; 0 < ||\mathbf{v}_{i}^{t}|| < \epsilon_{v}, \\
- \mu \; || (\mathbf{F}^{\mathrm{con}}_{i})^{n} || \; \frac {\mathbf{v}_{i}^{t}}  {||\mathbf{v}_{i}^{t}||} \; & \mathrm{when} \;  ||\mathbf{v}_{i}^{t}|| \geq \epsilon_{v},
\end{cases}
\label{eq:barrierFriction}
\end{equation}
where $\mu$ is the frictional coefficient and $\epsilon_{v}$ is the normalized parameter.
By introducing $\epsilon_{v}$, the traditional Heaviside frictional force can be smoothed and would be differentiable, such that the dynamic system can be solved variationally through a gradient-based method.

\paragraph{Fluid drag force} For the underwater locomotion, we need to consider the external drag force due to the existence of the fluids.
The drag force from the aqueous environments can be calculated as~\cite{mallick2014study,modarres2005nonlinear},
\begin{equation}
\mathbf{F}^{\mathrm{flu}}_{i} = - \frac{1} {2} \rho_{f} w  {\Delta \bar {l}}_{i} ( C_{d}^{\parallel} \mathbf{v}_{i}^{\parallel} || \mathbf{v}_{i}^{\parallel} || +  C_{d}^{\perp} \mathbf{v}_{i}^{\perp} || \mathbf{v}_{i}^{\perp} || ),
\end{equation}
where $\rho_f$ is the density of the fluid medium, $w$ is the width of the bilayers, $C_{d}^{\parallel}$ and $C_{d}^{\perp}$ are the drag coefficient along rod axial direction and normal direction, respectively, $\mathbf{v}_{i}$ is the velocity at $i$-th node, and $\mathbf{v}_{i}^{\parallel}$ (and $\mathbf{v}_{i}^{\perp}$) is its component along the segment tangent (and normal) direction.
This formulation is also smooth and differentiable for the implicit dynamic simulation.

\paragraph{Simulation step} Finally, considering the inertial effect, we formulate the dynamic equations of motion.
The diagonal mass matrix is denoted as $\mathbb{M} \in \mathbb{R}^{\mathcal{N} \times \mathcal{N}}$, thus the total kinetic energy is 
\begin{equation}
\mathcal{T} = \frac{1} {2} \dot{\mathbf{q}}^{T} \mathbb{M} \dot{\mathbf{q}}.
\end{equation}
The discrete equations of motion can be derived based on variational approach.
At the $k$-th time step, $t_{k}$, with the DOF vector, $\mathbf{q}(t_{k})$, and its velocity, $\dot{\mathbf{q}}(t_{k})$ at hand, the equations of motion and DOF vector can be updated from $t=t_{k}$ to $t=t_{k+1}$~\cite{Huang2019},
\begin{equation}
\begin{aligned}
{\mathbb{M}}  { \ddot{\mathbf{q}}(t_{k+1})} &- \mathbf{F}^{\text{ela}}(t_{k+1}) - \mathbf{F}^{\text{oth}}(t_{k+1}) = \mathbf{0}, \\
\mathbf{q}(t_{k+1}) &=  \mathbf{q}(t_{k}) + \dot{\mathbf{q}}(t_{k+1}) \; \delta t, \\
\dot{\mathbf{q}}(t_{k+1}) &=  \dot{\mathbf{q}}(t_{k}) + \ddot{\mathbf{q}}(t_{k+1}) \; \delta t,
\end{aligned}
\label{eq:implicitEuler}
\end{equation}
where $\mathbf{F}^{\text{ela}} = - \partial \mathcal{E} / \partial \mathbf{q}$ is the internal elastic force vector, $\mathbf{F}^{\text{oth}}$ is the other force vector (e.g.,  gravity, contact, and drag force), and $\delta t$ is the time step size.
The iterative Newton-Raphson method is employed to solve the nonlinear equations of motion.

\section{Model Validation}
\label{sec:validation}

To validate the accuracy, robustness, and versatility of our numerical framework, we first benchmark its performance against classical bilayer models. 
These benchmarks serve to verify the proposed framework's ability to simulate the fundamental mechanics of bilayer structures under various strain mismatches.

\subsection{2D validation: bending deformation}

As shown in Fig.~\ref{fig:validate}A, we consider a bilayer structure composed of an top layer and a bottom layer. 
The relative material properties are defined by $E_{2} = m E_{1}$ and $h_{2} = n h_{1}$, where $m$ and $n$ represent the ratios of Young's modulus and thickness of the both layers, respectively. 
When the top layer expands from an initial length $L$ to $L + \Delta L$, the strain mismatch between the two layers causes the bilayer structure to bend into an arc with curvature $\kappa$.
The normalized bending curvature $\kappa h / \eta$ can be predicted by the classical Timoshenko model~\cite{Timoshenko1925}, given as:
\begin{equation} 
\frac{\kappa h}{\eta} = \frac{6(1+m)^2}{3(1+m)^2 + (1+mn)(m^2+{1}/{mn})},
\label{eq:timoshenko} 
\end{equation}
where $\eta = \Delta L / L$ represents the expansion strain, and $h = h_1 + h_2$ denotes the total thickness of the bilayer. 
This widely used analytical formula provides a reliable theoretical benchmark for validating bilayer deformation models.

\begin{figure}[ht!]
  \centering  \includegraphics[width=1.0\linewidth]{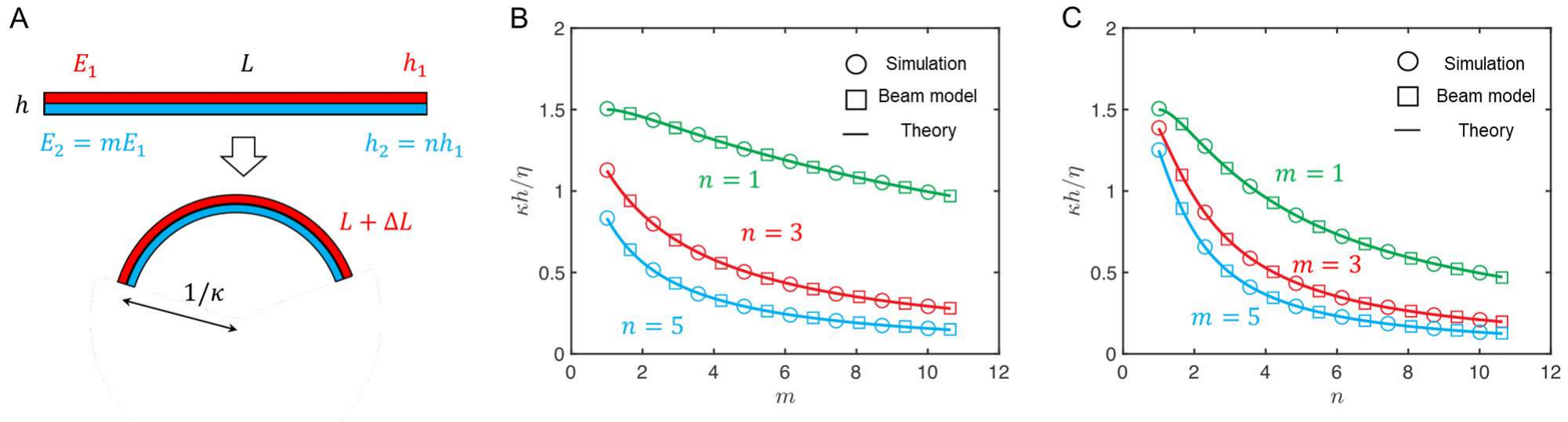}
  \caption{Model validation of bilayer structure. (A) Schematic diagram illustrating planar bending behavior of bilayer structure. The top and bottom layers are configured with Young's modulus $E_1$ and $E_2$, and thickness $h_1$ and $h_2$, respectively. The ratios $m$ and $n$ represent Young's modulus and thickness of the bottom layer relative to the top layer, respectively. (B) The normalized bending curvature $\kappa h / \eta$ plotted against Young's modulus ratio $m$. (C) The normalized bending curvature $\kappa h / \eta$ plotted against thickness ratio $n$. Comparisons are highlighted among the simulation results (circles), the simplified beam model theory (squares), and Timoshenko theory (solid line) as expressed in Eq.~(\ref{eq:timoshenko}).}
  \label{fig:validate}
\end{figure}

To evaluate the accuracy of our numerical framework, in Fig.~\ref{fig:validate}B and Fig.~\ref{fig:validate}C,  we compare the normalized bending curvatures $\kappa h / \eta$ obtained from simulations with predictions from a simplified 2D beam model and Eq.~(\ref{eq:timoshenko}) under varying Young's modulus ratios $m$ and thickness ratios $n$. 
These comparisons are illustrated in Fig.~\ref{fig:validate}B and Fig.~\ref{fig:validate}C, respectively. 
Across all tested cases, our numerical results (circles) align closely with both the simplified beam model(squares) and Timoshenko model (solid line). 
This strong agreement highlights the accuracy of our framework in capturing the essential mechanics of 2D bilayer bending.

\subsection{3D validation: helical deformation}
Beyond planar bending, bilayer structures are often subjected to more complex 3D deformations, such as bending combined with torsion. 
To validate the capability of our framework to handle such scenarios, we simulate the helical deformation of bilayer structures, which serves as a canonical example of 3D strain mismatch.

As shown in Fig.~\ref{fig:helix}, we consider a bilayer structure in which the top layer has an intrinsic helical curvature in its natural configuration, while the bottom layer remains straight. 
Upon bonding of the two layers at their interface, the resulting bilayer deforms into a helical shape with a pitch reduced compared to that of the natural state of the top layer. 
This deformation arises from the competition between the intrinsic curvature of the upper layer and the resistance of the lower layer.

The simulation successfully captures the coupling between bending and torsion, showcasing the flexibility of our framework in modeling 3D morphologies driven by mismatched strains. 
Such 3D validations are critical, as bilayer soft robots often rely on combined bending and twisting deformations for advanced functionalities, such as gripping, locomotion, and shape morphing.

\begin{figure}[ht!]
  \centering  \includegraphics[width=0.7\linewidth]{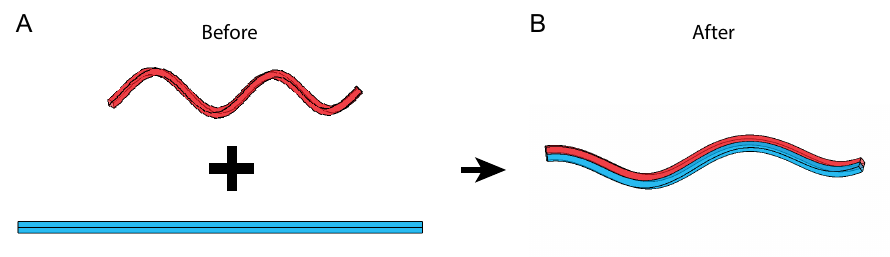}
  \caption{(A) The state of bilayer structures before deformation, in which the top layer has a natural configuration of a spiral helix and the bottom layer has a natural configuration of a straight line. (B) The state of the bilayer structure after deformation, in which the bilayer structure morphs into a complex state.}
  \label{fig:helix}
\end{figure}

The results of both 2D and 3D validation scenarios demonstrate the reliability and versatility of our numerical model. It accurately reproduces classical analytical predictions and effectively simulates complex, coupled deformations. This strong foundation establishes confidence in using the framework for more advanced and biologically inspired bilayer designs in subsequent studies.

\section{Robotic Demonstration}
\label{sec:demo}
In this section, we leverage the numerical framework to simulate various scenarios inspired by previous bilayer robotics studies, including gripping, crawling, jumping, and swimming. These simulations showcase the effectiveness of the DDG-based method in accurately modeling the behavior of bilayer soft robots.

\subsection{Gripping}

Bilayer structures can be leveraged to create soft robotic grippers capable of grasping objects in response to external stimuli~\cite{wang2019soft}.
To grasp a spherical object, two bilayer structures are joined at their midpoints, forming a perpendicular configuration. 
Upon gradual application of an external stimulus over 13.30 seconds, the bilayer gripper undergoes a controlled deformation, transitioning from an initially open state to a fully enclosed configuration, as illustrated in Fig.~\ref{fig:case1}.

\begin{figure}[ht!]
    \centering
    \includegraphics[width=1\textwidth]{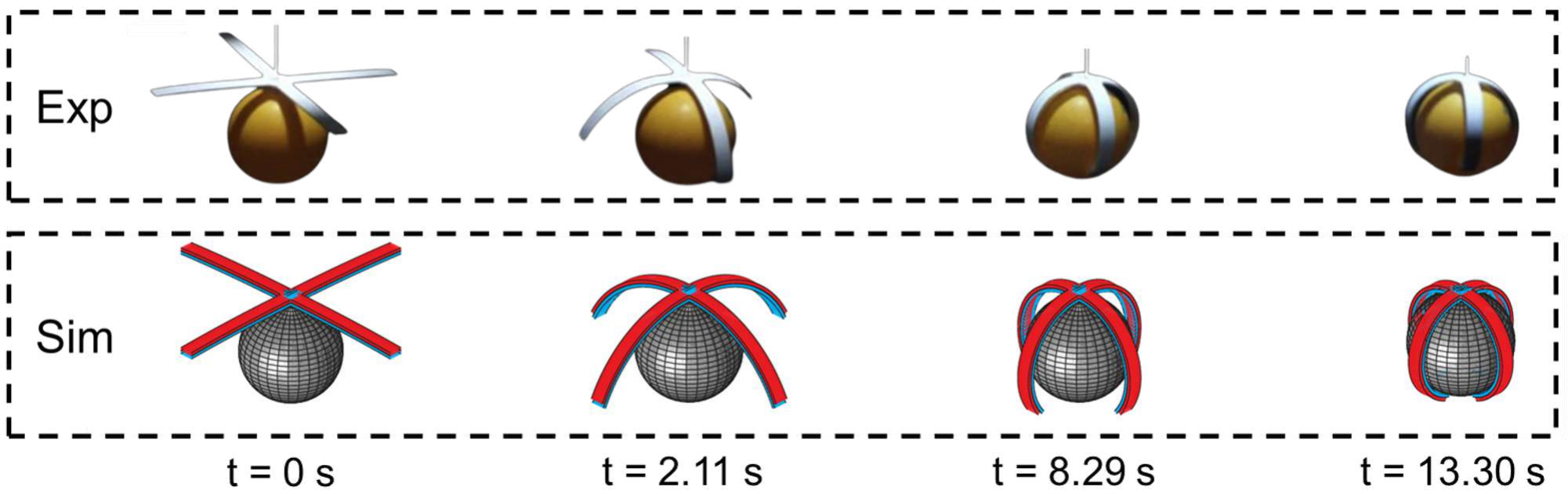}
    \caption{Gripping demonstration using a bilayer gripper to grasp a ping pong ball.  A sequence of snapshots illustrate the dynamic grasping process in both the experiment~\cite{wang2019soft} and the simulation. The gripper transitions over time from its initial straight configuration to a fully wrapped state, showcasing its ability to conform to the object's shape.} 
    \label{fig:case1}
\end{figure}

This dynamic grasping behavior arises from strain mismatch caused by the differential expansion of the bilayer components. 
Specifically, the reference length of the top layer increases over time, inducing bending due to the resulting strain mismatch. 
Once the curvature of the gripper surpasses a critical threshold, physical contact occurs between the soft gripper and the spherical object, enabling successful capture.
A strong quantitative agreement is observed between experimental results and numerical predictions, as shown in Fig.~\ref{fig:case1}. 
The complete dynamic process is available in Supporting Information Movie-S1.

\begin{figure}[ht!]
    \centering
    \includegraphics[width=1\textwidth]{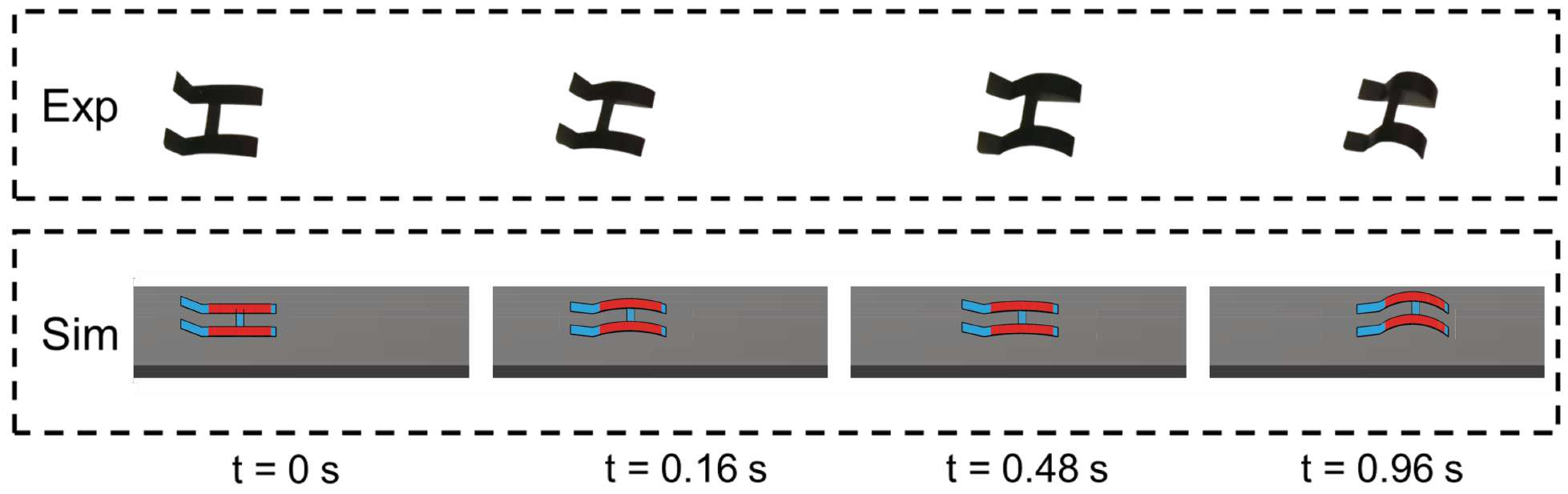}
    \caption{
    Crawling demonstration using a inchworm-like robot. A sequence of snapshots
    illustrate the dynamic crawling process in both the experiment~\cite{dong2019multi} and the simulation. The robot is actuated by IR light and moves through a two-phase motion cycle: the front feet grip the branch tightly while the rear section advances by curling its body; the rear feet then secure the branch as the front section extends forward, completing the crawling motion.
    } 
    \label{fig:case2}
\end{figure}

\subsection{Crawling}

Bilayer structures are widely employed in the design and fabrication of crawling soft robots.
To validate our simulation method, we analyzed two types of crawling robot: inchworm-like crawling robot~\cite{dong2019multi} and dual-leg crawling robot~\cite{shin2018hygrobot}, based on prior studies.

Inchworm-like crawling robot features a simple bilayer ribbon with tail-like extensions at its ends~\cite{dong2019multi}. 
The robot is usually made of the nanocellulose materials~\cite{li2024strain}, which can be actuated by the periodic changes in environmental moisture, causing the top layer to alternately swell and shrink. 
This hydration-driven deformation enables the robot to crawl forward.
Our simulation results for the inchworm-like crawling robot are shown in Fig.~\ref{fig:case2}.
We applied different friction coefficients for forward motion ($\mu=0.3$) and backward motion ($\mu=0.6$) to simulate directional movement. 
This frictional asymmetry is consistent with experimental observations~\cite{dong2019multi}.

While the inchworm-like crawling robot achieves locomotion under alternating hydration stimuli, its crawling velocity can be significantly improved by incorporating a dual-leg design~\cite{shin2018hygrobot}.
As shown in Fig.~\ref{fig:case3}, The dual-leg crawling robot enhances efficiency by leveraging two key factors:
first, reduced friction, as the legs decrease the robot’s contact area with the ground, minimizing overall resistance; 
second, increased backward displacement, where the legs extend the effective distance during the backward stroke when the body bends to a specific curvature, amplifying forward motion. 

\begin{figure}[ht!]
    \centering
    \includegraphics[width=1\textwidth]{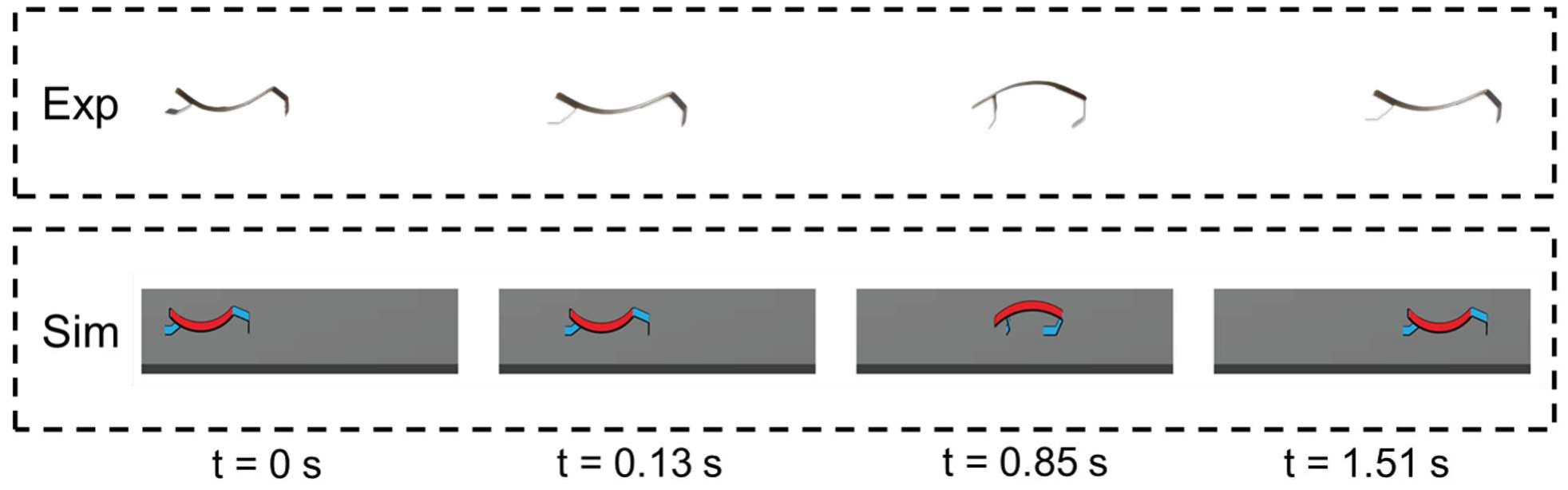}
    \caption{
     Crawling demonstration using a dual-leg robot. A sequence of snapshots illustrate the dynamic crawling process in both the experiment~\cite{shin2018hygrobot} and the simulation. The robot is actuated by the humidity difference. When humidity is high, bilayer bending pulls the legs together, moving only the hind leg due to its lower friction. As the bilayer dries and stretches out, the legs are pushed apart, causing only the foreleg to move, which has less friction than the hind leg.
    } 
    \label{fig:case3}
\end{figure}

Quantitative agreement between experiments and numerical predictions is demonstrated in Fig.~\ref{fig:case2} and Fig.~\ref{fig:case3}. 
The full dynamic process of the crawling robots can be found in Supporting Information Movie-S1.

\subsection{Jumping}

Bilayer soft robots demonstrate remarkable jumping capabilities, enabling rapid locomotion and obstacle traversal in unstructured environments~\cite{li2022dual,xu2022insect,hu2017electrically}.
These robots leverage a unique actuation mechanism that combines an encapsulated shape memory alloy (SMA) wire with an elastomeric bilayer, allowing for controlled energy storage and release.
Upon electrical activation, the SMA wire undergoes rapid contraction due to its thermally induced phase transition, generating a strong bending deformation in the bilayer structure.
This deformation not only modifies the shape of the robot but also plays a crucial role in accumulating elastic potential energy within the bilayer.
Once the actuation stops and the SMA wire returns to its original state, the stored energy is abruptly released, propelling the robot into a dynamic jumping motion. 
Fig.~\ref{fig:case4} illustrates the complete jumping sequence in both simulation and experiment, showcasing strong agreement between numerical predictions and experimental observations.
A detailed visualization of the dynamic jumping process is available in Supporting Information Movie-S1.

\begin{figure}[ht!]
    \centering
    \includegraphics[width=1\textwidth]{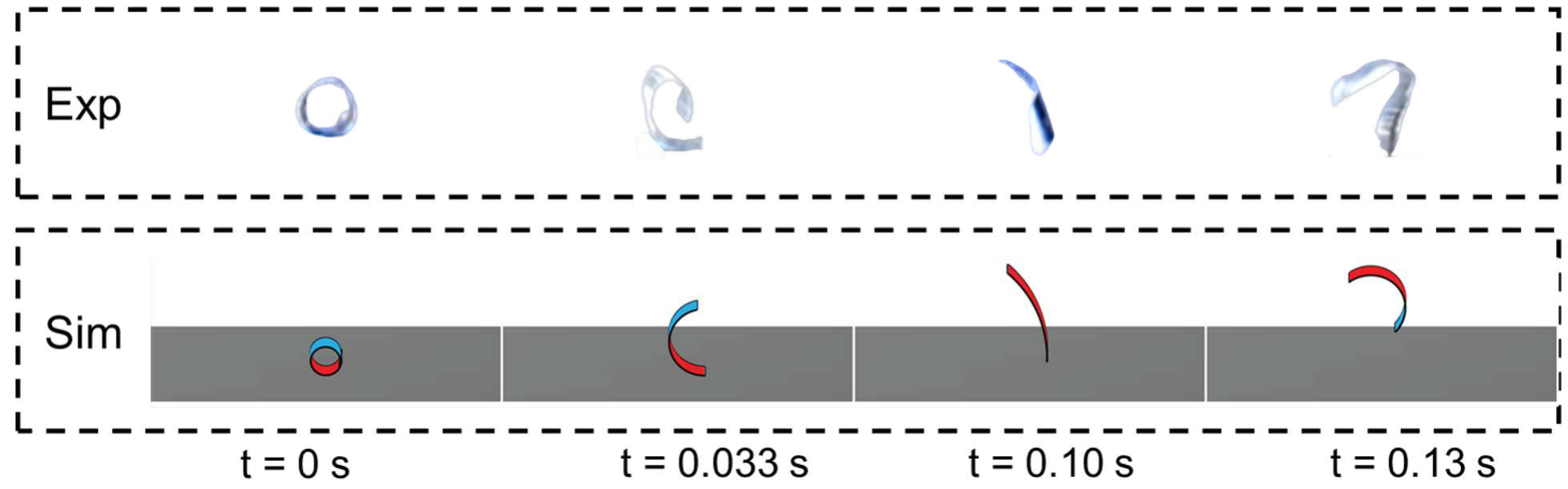}
    \caption{
    Jumping demonstration using an insect larvae-inspired robot. A sequence of snapshots illustrate the dynamic jumping process in both the experiment~\cite{xu2022insect} and the simulation. The robot is actuated by light irradiation, which triggers a delayed yet sudden jump. When exposed to light from the upper left, the actuator remains stationary for a brief period before rapidly releasing stored elastic energy, propelling the robot vertically.} 
    \label{fig:case4}
\end{figure}

\subsection{Swimming}

Beyond gripping, crawling, and jumping, bilayer soft robots also exhibit swimming capabilities in aqueous environments when exposed to external light stimuli~\cite{yin2021visible,huang2022design}.
In this section, we utilize our developed simulation framework to investigate the swimming dynamics of these robots. 
By incorporating implicit drag forces, computed using Eq.~(\ref{eq:barrierContact})~\cite{mallick2014study,modarres2005nonlinear}, we accurately model the interaction between the robot and the surrounding fluid. 
The propulsion mechanism arises from the applied periodic light stimuli between the loading and unloading processes, enabling the soft robot to generate thrust and swim forward.
Because of the periodic bending actuated by the light stimuli, the soft robots can utilize the propulsion force to swim forward.
Fig.~\ref{fig:case5} presents a sequence of snapshots comparing the swimming motion of the bilayer soft robot in simulation and experiments, demonstrating a strong agreement between the two. 
A dynamic rendering of the swimming process can be found in Supporting Information Movie-S1.

\begin{figure}[ht!]
    \centering
    \includegraphics[width=1\textwidth]{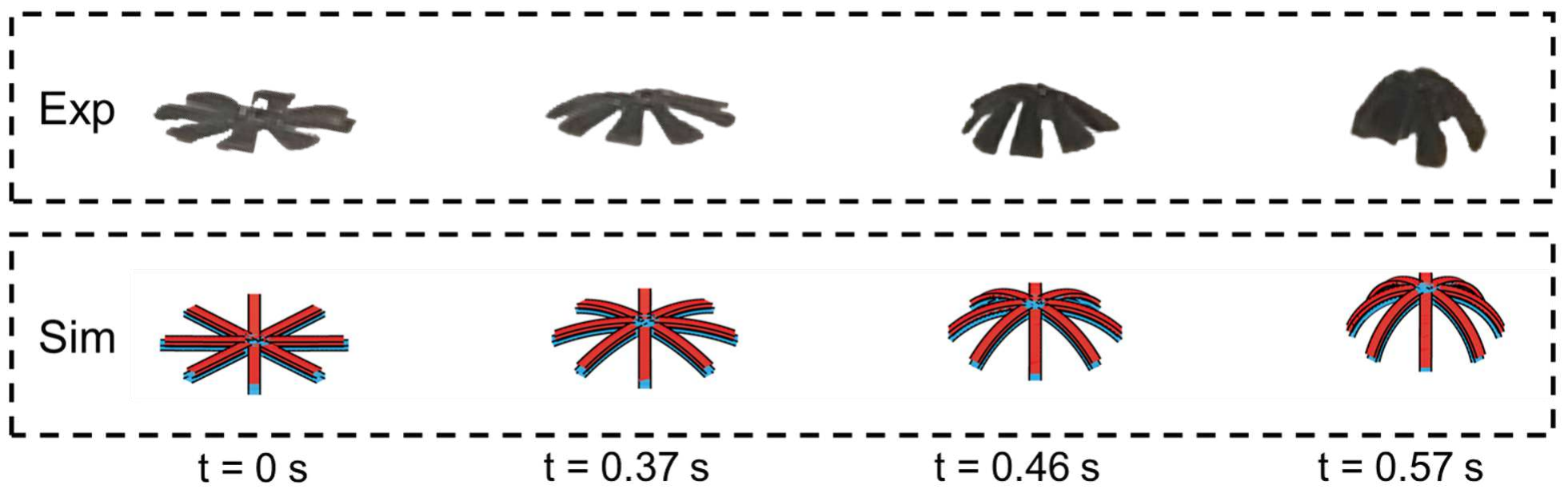}
    \caption{
    Swimming demonstration using an jellyfish-like robot. A sequence of snapshots illustrate the dynamic swimming process in both the experiment~\cite{yin2021visible} and the simulation. The robot is actuated by light, and as it deforms under light exposure, the surrounding fluid flows inward toward the center, propelling the robot upward. When the light is turned off, the robot gradually recovers from the bending deformation, and the surrounding fluid velocity progressively decreases.
    } 
    \label{fig:case5}
\end{figure}

\section{Conclusion}
 \label{sec:conclusion}
In this paper, we present the development of a simulation environment based on the Discrete Elastic Rod (DER) framework, tailored for the dynamic analysis of bilayer soft robots. 
This environment is constructed by implementing the DER model and formulating the constraints governing the rigid body motion at the bilayer interface. 
The simulation accuracy is validated through comparisons with established classical benchmarks, ensuring reliability and precision.
To highlight the capabilities of the environment in addressing complex interactions, we simulate a variety of real-world robotic scenarios. 
We integrate a frictional contact model and a fluid-structure interaction model within the numerical framework, enabling the simulation of diverse dynamic behaviors such as gripping, crawling, jumping, and swimming.
This simulation platform offers powerful tools for analyzing the dynamic performance of bilayer soft robots, particularly in environments where environmental interactions play a crucial role. 
It serves as a robust foundation for model-based design and control strategies, paving the way for advanced applications of these versatile soft robotic systems.

\section*{Supporting Information} \par 
We provide a video as supporting information to show the dynamic behaviors of the bilayer soft robots.

\section*{Appendix A: Discrete bilayer beam model}

In this section, we present a discrete model for the analysis of a bilayer beam.
Instead of discretizing the two layers' centerline separately and then employing the coupling element at the interface, here we use the interface as the DOF vector to represent the configuration of the bilayer beam. 
The interface is discretized into $N$ nodes and, as the out-of-plane deformation is ignored, we only consider the stretching v and the in-plane bending mode. 
The strain for each stretching element and bending element is identical to the 3D rod case, and we use $\epsilon^{i}$ to represent the stretching strain of $i$-th edge and use $\kappa_{i}$ to represent the planar curvature of $i$-th node.
The intrinsic strain associated with the top layer at $i$-th node is defined as $\eta_{i}$.
The total elastic energy of $i$-th node is
\begin{equation}
\begin{aligned}
\mathcal{E}_{i}^{\mathrm{2D}} &= \int_{0}^{h1}\frac {1} {2} E_{1} w \Delta l_{i}  ( \epsilon_{i} -\eta_{i} + \kappa_{i} z )^2 dz + \int_{-h2}^{0}  \frac {1} {2} E_{2} w \Delta l_{i}  ( \epsilon_{i} + \kappa_{i} z )^2 dz  \\
&= \frac{1} {6}  E_{1} w h_{1} \left( 3 \eta_{i}^2 + h_{1}^2\kappa_{i}^2 - 3  h_{1} \eta_{i} \kappa_{i} - 6 \eta_{i} \epsilon_{i} + 3 h_{1}\kappa_{i}  \epsilon_{i} + 3 \epsilon_{i}^2  \right)\Delta l_{i} + \frac{1} {6}  E_{2} w h_{2} \left( h_{2}^2 \kappa_{i}^2 - 3h_{2} \epsilon_{i} \kappa_{i} +  3 \epsilon_{i}^2 \right) \Delta l_{i},
\end{aligned}
\end{equation}
where $\epsilon_{i}  = ( \epsilon^{i-1} + \epsilon^{i} ) / 2$ is the stretching strain associated with $i$-th node.
The total elastic energy for this case is
\begin{equation}
\mathcal{E}^{\mathrm{2D}} =  \sum_{i=1}^{N-2} \mathcal{E}_{i}^{\mathrm{2D}}.
 \end{equation}
A similar procedure is used to solve the dynamic equilibrium for the simplified 2D beam system.
In this simulation, we gradually modify $\eta_{i}$ to achieve the strain change of the top layer.

\section*{Appendix B: The physical and geometric parameters used in simulations}

Here, we list the geometric and physical parameters used in our numerical study, as shown in Table 1.

\begin{table}[h!]
\centering
\caption{Physical and geometric parameters}\label{tableData}
\begin{tabular}{c|c|c|c|c|c|c|c|c}
Parameter & Notation & Unit & Gripping & Crawling 1 & Crawling 2 & Jumping & Swimming \\ \hline
Top layer stiffness & $E_{1}$ & $\mathrm{MPa}$ &  $100$ & $100$ & $100$ & $100$ & $100$ &\\
Bottom layer stiffness & $E_{2}$ & $\mathrm{MPa}$ &  $100$ & $100$ & $100$ & $100$ & $100$ & \\
Top layer thickness & $h_{1}$ & $\mathrm{mm}$ & $1.0$ & $1.0$ & $1.0$ & $1.0$ &$1.0$ &\\
Bottom layer thickness & $h_{2}$ & $\mathrm{mm}$ & $1.0$ & $1.0$ & $1.0$ & $1.0$ &$1.0$ &\\
Gravity & $g$ & $\mathrm{m/s^2}$ & $-10.0$ & $-10.0$ & $-10.0$ & $-10.0$ &$0.0$ &\\
Penalty stiffness & $K_{T} / EA$ & - &   $100$ & $100$ & $100$ & $100$ & $100$ &\\
Penalty stiffness & $K_{R} / EI$ & - &  $1000$ &  $1000$ &  $1000$ &  $1000$ &  $1000$ &\\
Ribbon width & $w$ & $\mathrm{mm}$ & $10$ & $10$ & $10$ & $10$ &$10$ &\\
Material density & $\rho$ & $\mathrm{kg/m^3}$ & $1000$ & $1000$ & $1000$ &$1000$ & $1000$& \\
Top layer nodes & $M$ & $\mathrm{-}$ & $201$ & $80$ & $100$ &$100$ &$321$ & \\
Bottom layer nodes & $N$ & $\mathrm{-}$ & $192$ & $50$ & $55$ &$100$ &$288$ & \\
Time step size & $\delta t$ & $\mathrm{ms}$ & $1.0$ & $1.0$ & $1.0$ &$1.0$ &$1.0$ & \\
Contact barrier & $\tilde{d}$ & $\mathrm{m}$ & $0.001$ & $0.001$ & $0.001$ &$0.001$ & -- & \\
Contact stiffness & $K_{c}$ & $\mathrm{MPa}$ & $10$ & $10$ & $10$ &$10$ & - & \\ 
Velocity barrier & $\epsilon_{v}$ & $\mathrm{m/s}$ &  - & $0.0001$ & $0.0001$ &$0.0001$ & - & \\
Friction coefficient & $\mu$ & - &  - & $\{0.6,0.3\}$ & $\{0.6,0.3\}$ & $0.5$ & - & \\
Drag coefficient & $C_{d}^{\parallel}$ & - & - & - &  - & - & 0.01 & \\
Drag coefficient & $C_{d}^{\perp}$  & - & - & - &  - & - & 1.0 & \\
\end{tabular}
\end{table}

\medskip
\textbf{Acknowledgements} \par 

M.L. acknowledges the start-up funding from the University of Birmingham.
W.H. acknowledges the start-up funding from Newcastle University, UK.

\medskip

%
\bibliographystyle{MSP}
\bibliography{paper}

\providecommand{\noopsort}[1]{}\providecommand{\singleletter}[1]{#1}%
\begin{thebibliography}{10}
\providecommand{\url}[1]{\texttt{#1}}
\providecommand{\urlprefix}{URL }

\bibitem{wang2019soft}
W.~Wang, C.~Y. Yu, P.~A.~A. Serrano, S.-H. Ahn,
\newblock \emph{Composites Part B: Engineering} \textbf{2019}, \emph{164} 198.

\bibitem{shin2018hygrobot}
B.~Shin, J.~Ha, M.~Lee, K.~Park, G.~H. Park, T.~H. Choi, K.-J. Cho, H.-Y. Kim,
\newblock \emph{Science Robotics} \textbf{2018}, \emph{3}, 14 eaar2629.

\bibitem{li2022dual}
J.~Li, M.~Wang, Z.~Cui, S.~Liu, D.~Feng, G.~Mei, R.~Zhang, B.~An, D.~Qian,
  X.~Zhou, et~al.,
\newblock \emph{Journal of Materials Chemistry A} \textbf{2022}, \emph{10}, 47
  25337.

\bibitem{xu2022insect}
L.~Xu, F.~Xue, H.~Zheng, Q.~Ji, C.~Qiu, Z.~Chen, X.~Zhao, P.~Li, Y.~Hu,
  Q.~Peng, et~al.,
\newblock \emph{Nano Energy} \textbf{2022}, \emph{103} 107848.

\bibitem{hu2017electrically}
Y.~Hu, J.~Liu, L.~Chang, L.~Yang, A.~Xu, K.~Qi, P.~Lu, G.~Wu, W.~Chen, Y.~Wu,
\newblock \emph{Advanced Functional Materials} \textbf{2017}, \emph{27}, 44
  1704388.

\bibitem{yin2021visible}
C.~Yin, F.~Wei, S.~Fu, Z.~Zhai, Z.~Ge, L.~Yao, M.~Jiang, M.~Liu,
\newblock \emph{ACS Applied Materials \& Interfaces} \textbf{2021}, \emph{13},
  39 47147.

\bibitem{huang2022design}
X.~Huang, Z.~J. Patterson, A.~P. Sabelhaus, W.~Huang, K.~Chin, Z.~Ren, M.~K.
  Jawed, C.~Majidi,
\newblock \emph{Advanced Intelligent Systems} \textbf{2022}, \emph{4}, 10
  2200163.

\bibitem{tao2021morphing}
Y.~Tao, Y.-C. Lee, H.~Liu, X.~Zhang, J.~Cui, C.~Mondoa, M.~Babaei,
  J.~Santillan, G.~Wang, D.~Luo, et~al.,
\newblock \emph{Science Advances} \textbf{2021}, \emph{7}, 19 eabf4098.

\bibitem{kim2022magnetic}
Y.~Kim, X.~Zhao,
\newblock \emph{Chemical Reviews} \textbf{2022}, \emph{122}, 5 5317.

\bibitem{yang2023morphing}
X.~Yang, Y.~Zhou, H.~Zhao, W.~Huang, Y.~Wang, K.~J. Hsia, M.~Liu,
\newblock \emph{Soft Science} \textbf{2023}, \emph{3}, 4 38.

\bibitem{armon2011geometry}
S.~Armon, E.~Efrati, R.~Kupferman, E.~Sharon,
\newblock \emph{Science} \textbf{2011}, \emph{333}, 6050 1726.

\bibitem{ChenZi2010}
Z.~Chen, C.~Majidi, D.~J. Srolovitz, M.~Haataja,
\newblock \emph{Applied Physics Letters} \textbf{2011}, \emph{98}, 1 011906.

\bibitem{lijh2025}
J.~Li, X.~Sun, Z.~He, Y.~Hou, H.~Wu, Y.~Zhu,
\newblock \emph{Journal of the Mechanics and Physics of Solids} \textbf{2025},
  \emph{196} 105999.

\bibitem{Timoshenko1925}
S.~Timoshenko,
\newblock \emph{J. Opt. Soc. Am.} \textbf{1925}, \emph{11}, 3 233.

\bibitem{Efrati2009}
E.~Efrati, E.~Sharon, R.~Kupferman,
\newblock \emph{Journal of the Mechanics and Physics of Solids} \textbf{2009},
  \emph{57}, 4 762.

\bibitem{VanRees2017}
W.~M. van Rees, E.~Vouga, L.~Mahadevan,
\newblock \emph{Proceedings of the National Academy of Sciences} \textbf{2017},
  \emph{114}, 44 11597.

\bibitem{grinspun2006discrete}
E.~Grinspun, M.~Desbrun, K.~Polthier, P.~Schr{\"o}der, A.~Stern,
\newblock \emph{ACM Siggraph Course} \textbf{2006}, \emph{7}, 1.

\bibitem{tong2023snap}
D.~Tong, A.~Choi, J.~Joo, A.~Borum, M.~Khalid~Jawed,
\newblock \emph{Journal of Applied Mechanics} \textbf{2023}, \emph{90}, 4
  041008.

\bibitem{tong2021automated}
D.~Tong, A.~Borum, M.~K. Jawed,
\newblock \emph{IEEE Robotics and Automation Letters} \textbf{2021}, \emph{7},
  2 1126.

\bibitem{choi2021implicit}
A.~Choi, D.~Tong, M.~K. Jawed, J.~Joo,
\newblock \emph{Journal of Applied Mechanics} \textbf{2021}, \emph{88}, 5
  051010.

\bibitem{tong2023fully}
D.~Tong, A.~Choi, J.~Joo, M.~K. Jawed,
\newblock \emph{Extreme Mechanics Letters} \textbf{2023}, \emph{58} 101924.

\bibitem{choi2023deep}
A.~Choi, D.~Tong, D.~Terzopoulos, J.~Joo, M.~K. Jawed,
\newblock \emph{arXiv preprint arXiv:2301.01968} \textbf{2023}.

\bibitem{tong2024sim2real}
D.~Tong, A.~Choi, L.~Qin, W.~Huang, J.~Joo, M.~K. Jawed,
\newblock \emph{The International Journal of Robotics Research} \textbf{2024},
  \emph{43}, 6 791.

\bibitem{huang2020dynamic}
W.~Huang, X.~Huang, C.~Majidi, M.~K. Jawed,
\newblock \emph{Nature communications} \textbf{2020}, \emph{11}, 1 2233.

\bibitem{huang2023modeling}
W.~Huang, M.~Liu, K.~J. Hsia,
\newblock \emph{Extreme Mechanics Letters} \textbf{2023}, \emph{59} 101967.

\bibitem{huang2020numerical}
W.~Huang, M.~Jawed,
\newblock \emph{Soft matter} \textbf{2020}, \emph{16}, 3 604.

\bibitem{tong2024inverse}
D.~Tong, Z.~Hao, M.~Liu, W.~Huang,
\newblock \emph{IEEE Robotics and Automation Letters} \textbf{2024}.

\bibitem{huang2024dynamic}
W.~Huang, P.~Xu, Z.~Liu,
\newblock \emph{Journal of Applied Mechanics} \textbf{2024}, 1--13.

\bibitem{choi2024dismech}
A.~Choi, R.~Jing, A.~Sabelhaus, M.~K. Jawed,
\newblock \emph{IEEE Robotics and Automation Letters} \textbf{2024}.

\bibitem{Khalid2008}
M.~K. Jawed, A.~Novelia, O.~M. O'Reilly,
\newblock \emph{A primer on the kinematics of discrete elastic rods},
\newblock Springer, \textbf{2018}.

\bibitem{BergouDVT}
M.~Bergou, B.~Audoly, E.~Vouga, M.~Wardetzky, E.~Grinspun,
\newblock \emph{ACM Trans. Graph.} \textbf{2010}, \emph{29}, 4.

\bibitem{bergou2008discrete}
M.~Bergou, M.~Wardetzky, S.~Robinson, B.~Audoly, E.~Grinspun,
\newblock In \emph{ACM SIGGRAPH 2008 papers}, 1--12. ACM, \textbf{2008}.

\bibitem{Huang2019}
W.~Huang, M.~K. Jawed,
\newblock \emph{Journal of Applied Mechanics} \textbf{2019}, \emph{86}, 8
  084501.

\bibitem{audoly2000elasticity}
B.~Audoly, Y.~Pomeau,
\newblock In \emph{Peyresq Lectures on Nonlinear Phenomena}, 1--35. World
  Scientific, \textbf{2000}.

\bibitem{jawed2018primer}
M.~K. Jawed, A.~Novelia, O.~M. O'Reilly,
\newblock \emph{A primer on the kinematics of discrete elastic rods},
\newblock Springer, \textbf{2018}.

\bibitem{li2020incremental}
M.~Li, Z.~Ferguson, T.~Schneider, T.~R. Langlois, D.~Zorin, D.~Panozzo,
  C.~Jiang, D.~M. Kaufman,
\newblock \emph{ACM Trans. Graph.} \textbf{2020}, \emph{39}, 4 49.

\bibitem{li2020codimensional}
M.~Li, D.~M. Kaufman, C.~Jiang,
\newblock \emph{arXiv preprint arXiv:2012.04457} \textbf{2020}.

\bibitem{mallick2014study}
M.~Mallick, A.~Kumar, N.~Tamboli, A.~Kulkarni, P.~Sati, V.~Devi, S.~Chandar,
\newblock \emph{International Journal of Civil Engineering Research}
  \textbf{2014}, \emph{5}, 4 301.

\bibitem{modarres2005nonlinear}
Y.~Modarres-Sadeghi, M.~Pa{\"\i}doussis, C.~Semler,
\newblock \emph{Journal of Fluids and Structures} \textbf{2005}, \emph{21}, 5-7
  609.

\bibitem{dong2019multi}
Y.~Dong, J.~Wang, X.~Guo, S.~Yang, M.~O. Ozen, P.~Chen, X.~Liu, W.~Du, F.~Xiao,
  U.~Demirci, et~al.,
\newblock \emph{Nature Communications} \textbf{2019}, \emph{10}, 1 4087.

\bibitem{li2024strain}
J.~Li, Y.~Hou, Z.~He, H.~Wu, Y.~Zhu,
\newblock \emph{Nano Letters} \textbf{2024}.

\end{thebibliography}
\end{justify}





\end{document}